\pdfoutput=1

\documentclass[11pt]{article}

\usepackage[final]{acl}

\usepackage{times}
\usepackage{latexsym}

\usepackage[T1]{fontenc}

\usepackage[utf8]{inputenc}

\usepackage{microtype}

\usepackage{inconsolata}

\usepackage{graphicx}

\title{On Psychology of AI -- Does Primacy Effect Affect ChatGPT and Other LLMs?}

\author{Mika Hämäläinen \\
  Metropolia University of Applied Sciences \\
  Helsinki, Finland \\
  \texttt{firstname.lastname@metropolia.fi} \\}

\begin{document}
\maketitle
\begin{abstract}
We study the primacy effect in three commercial LLMs: ChatGPT, Gemini and Claude. We do this by repurposing the famous experiment Asch (1946) conducted using human subjects. The experiment is simple, given two candidates with equal descriptions which one is preferred if one description has positive adjectives first before negative ones and another description has negative adjectives followed by positive ones. We test this in two experiments. In one experiment, LLMs are given both candidates simultaneously in the same prompt, and in another experiment, LLMs are given both candidates separately. We test all the models with 200 candidate pairs. We found that, in the first experiment, ChatGPT preferred the candidate with positive adjectives listed first, while Gemini preferred both equally often. Claude refused to make a choice. In the second experiment, ChatGPT and Claude were most likely to rank both candidates equally. In the case where they did not give an equal rating, both showed a clear preference to a candidate that had negative adjectives listed first. Gemini was most likely to prefer a candidate with negative adjectives listed first.
\end{abstract}

\section{Introduction}

Large language models (LLMs) are becoming increasingly human-like in many aspects such as language use \cite{PPR629825}, cognitive biases \cite{azaria2023chatgpt} and problem solving \cite{orru2023human}. This has led us to a world where LLMs are perhaps better studied from the perspective of humanities and psychology than through typical NLP benchmarks \cite{hamalainen2024growing}.

It is known that the order in which information is presented can have a profound impact on how it is perceived and interpreted, a phenomenon often referred to as the primacy effect \cite{asch1946forming}. For example, in one of Asch's (\citeyear{asch1946forming}) experiments, participants were asked to evaluate a person after being presented with a list of descriptive words. When these words progressed from high favorability to low favorability  or from low favorability to high favorability, participants consistently formed stronger impressions based on the information encountered earlier, highlighting the power of initial traits to anchor (see \citealt{furnham2011literature}) subsequent evaluations.

In other words, the primacy effect refers to the human tendency to give greater weight to early information in a sequence, shaping how subsequent details are interpreted. This bias has implications that extend beyond simple word lists, influencing social perception, decision-making and memory.

This paper will explore the primacy effect as defined by Asch’s (\citeyear{asch1946forming}) findings in three commercial LLMs: ChatGPT, Claude and Gemini. We conduct two experiments where we assess whether the LLMs show preference for one of two candidates with identical characteristics based on the orded in which the characteristics are presented. 

\section{Related Work}

\begin{table*}[ht]
\centering
\begin{tabular}{lll}
1. generous - ungenerous    & 7. popular - unpopular          & 13. serious - frivolous        \\
2. wise - shrewd            & 8. reliable - unreliable        & 14. talkative - restrained     \\
3. happy - unhappy          & 9. important - insignificant    & 15. altruistic - self-centered \\
4. good-natured - irritable & 10. humane - ruthless           & 16. imaginative - hard-headed  \\
5. humorous - humorless     & 11. good-looking - unattractive & 17  strong - weak              \\
6. sociable - unsociable    & 12. persistent - unstable       & 18. honest - dishonest        
\end{tabular}
\caption{The 18 antonym pairs used to build the dataset}
\label{tab:characteristics}
\end{table*}

\begin{table*}[ht]
\centering
\resizebox{\textwidth}{!}{%
\begin{tabular}{|l|l|l|}
\hline
A                                                                         & B                                                                         & Positive first \\ \hline
restrained - ungenerous - unreliable - humorous - strong - important      & humorous - strong - important - restrained - ungenerous - unreliable      & B              \\ \hline
sociable - good-natured - talkative - unstable - hard-headed - ungenerous & unstable - hard-headed - ungenerous - sociable - good-natured - talkative & A              \\ \hline
shrewd - unpopular - unsociable - generous - reliable - humorous          & generous - reliable - humorous - shrewd - unpopular - unsociable          & B              \\ \hline
wise - honest - good-natured - unstable - ungenerous - weak               & unstable - ungenerous - weak - wise - honest - good-natured               & A              \\ \hline
unsociable - shrewd - humorless - humane - good-looking - popular         & humane - good-looking - popular - unsociable - shrewd - humorless         & B              \\ \hline
popular - serious - generous - unsociable - insignificant - unhappy       & unsociable - insignificant - unhappy - popular - serious - generous       & A              \\ \hline
altruistic - good-looking - wise - unreliable - irritable - unsociable    & unreliable - irritable - unsociable - altruistic - good-looking - wise    & A              \\ \hline
\end{tabular}%
}
\caption{An example of the generated data}
\label{tab:example-data}
\end{table*}

Primacy effect is a well-studied phenomenon in the field of psychology \cite{anderson1961primacy,decoster2004meta}. In this, section we will focus on the recent NLP research on the topic.

A recent study \cite{wang-etal-2023-primacy} explores this issue by examining the primacy effect in ChatGPT, defined as the tendency to favor labels presented earlier in a sequence. The findings reveal two key points: (i) ChatGPT’s decisions are sensitive to the order of labels in the prompt, and (ii) it exhibits a significantly higher likelihood of selecting labels in earlier positions as answers. These insights highlight the potential for cognitive biases to emerge in LLM-based systems.

Another recent research paper \cite{guo2024serialpositioneffectslarge} suggests that LLMs may exhibit serial position effects, such as primacy and recency biases, which are well-documented cognitive phenomena in human psychology. Testing across a variety of labeling tasks and models confirms the prevalence of these effects, although their intensity varies depending on the context. Notably, while carefully designed prompts can help mitigate these biases to some extent, their effectiveness remains inconsistent.

Although there is recent NLP research on the very same topic, the prior research focuses on labeling tasks rather than a task that has been used to study human psychology. Our research will thus contribute through a new aspect of studying the primacy effect in LLMs.

\section{Data}

As we draw inspiration from Asch's (\citeyear{asch1946forming}) famous experiment by conducting a similar experiment in a computational setting in our Experiment 1, we use the word list presented in the original paper. The word list consists of pairs of antonyms, describing the a trait in a positive and negative way. This list of antonym pairs can be seen in Table \ref{tab:characteristics}.

In Asch's (\citeyear{asch1946forming}) study, participants were presented with two candidates who were described by six adjectives. For one candidate, the list of adjectives contained 3 positive adjectives followed by 3 negative ones. For the other candidate, the list of adjectives had the same adjectives but in an inverse order of polarity, that is 3 negative adjectives followed by 3 positive adjectives. In a similar fashion, we generate a dataset of 200 description pairs, both described by the same 6 adjectives but in a different order of polarity. As in the original study, 3 of the adjectives are positive and 3 negative. The positive and negative adjectives cannot be each other's antonyms as that would result in a contradictory description.

We pick the adjectives at random from the pool of candidate adjectives for each description pair. Each description pair has two candidates: Candidate A and Candidate B. Which candidate has the positive adjectives first is also picked at random. This way, our dataset has 3 columns, one for both descriptions of the candidates and one that indicates which candidate has the positive adjectives listed first in their description. An example of this data can be seen in Table \ref{tab:example-data}.

\begin{table*}[ht]
\centering
\resizebox{\textwidth}{!}{%
\begin{tabular}{|l|l|}
\hline
Prompt template                                                                                                                                                                                                                                                                                  & Example prompt                                                                                                                                                                                                                                                                                                                                                                                                     \\ \hline
\begin{tabular}[c]{@{}l@{}}I have two candidates, but I can only invite one to an interview. Based on the \\ characteristics of each candidate, help me decide which one to invite. \\ Please answer only candidate A or B.\\ \\ Candidate A: ADJECTIVES1\\ Candidate B: ADJECTIVES2\end{tabular} & \begin{tabular}[c]{@{}l@{}}I have two candidates, but I can only invite one to an interview. Based on the \\ characteristics of each candidate, help me decide which one to invite. \\ Please answer only candidate A or B.\\ \\ Candidate A: restrained - ungenerous - unreliable - humorous - strong - important\\ Candidate B: humorous - strong - important - restrained - ungenerous - unreliable\end{tabular} \\ \hline
\end{tabular}%
}
\caption{Prompt used in Experiment 1}
\label{tab:experiment1-prompts}
\end{table*}

\begin{table*}[ht]
\centering
\resizebox{\textwidth}{!}{%
\begin{tabular}{|l|l|}
\hline
Prompt template                                                                                                                                                                                                                                                                                                                                                                                     & Example prompt                                                                                                                                                                                                                                                                                                                                                                                                                                                \\ \hline
\begin{tabular}[c]{@{}l@{}}I am conducting a series of job interviews, and I have to decide whether I should \\ invite a candidate to an interview. Based on the following characteristics, rank \\ this candidate on a scale of 1-5. 1 meaning I should not interview them and 5 \\ meaning that I should interview them. Answer only with a number.\\ \\ Characteristics: ADJECTIVES\end{tabular} & \begin{tabular}[c]{@{}l@{}}I am conducting a series of job interviews, and I have to decide whether I should \\ invite a candidate to an interview. Based on the following characteristics, rank \\ this candidate on a scale of 1-5. 1 meaning I should not interview them and 5 \\ meaning that I should interview them. Answer only with a number.\\ \\ Characteristics: humorous - strong - important - restrained - ungenerous - unreliable\end{tabular} \\ \hline
\end{tabular}%
}
\caption{Prompt template and an example prompt for Experiment 2}
\label{tab:experiment2-prompt}
\end{table*}

\section{Experiment 1: Pick Between Two Candidates}

In the famous experiment by Asch (\citeyear{asch1946forming}), participants were shown descriptions of two identical candidates at a time with the only difference being the order in which the negative and positive adjectives appeared. In our first experiment, we will also give the LLMs descriptions of two identical candidates and ask the model to indicate its preference.

There are some key differences between the original study on human subjects and our study. First, Asch (\citeyear{asch1946forming}) never studied this phenomenon with as many different combinations of adjectival descriptions. In fact, they only report results on two different sets of adjectival descriptions.

Another key difference is that Asch (\citeyear{asch1946forming}) invited the test subjects to describe each candidate by using a fixed list of antonyms (same ones as in Table \ref{tab:characteristics}) and also to give a qualitative description of the candidates. Instead of this test setup, we ask the LLMs to pick either candidate A or B and respond only with the choice they made. We do this because we want to avoid inadvertently triggering a chain-of-thought reasoning in some of the LLM responses. Instead, we are interested in seeing what the implicit attitude is the LLM holds towards each candidate by requesting a rapid response.

The prompt template and an example prompt can be seen in Table \ref{tab:experiment1-prompts}. We send this template filled with the 200 test cases to each LLM over their respective APIs. The models that are in use are GPT-4o for ChatGPT, Gemini 1.5 Flash and Claude 3.5 Sonnet Latest. The experiment was conducted on the 20th of January in 2025.


\begin{table}[ht]
\centering
\resizebox{\columnwidth}{!}{%
\begin{tabular}{|l|l|l|l|}
\hline
                  & ChatGPT & Gemini & Claude \\ \hline
Positive first    & 65.5\%  & 47.5\% & 0\%    \\ \hline
Negative first    & 31\%    & 47.5\% & 0\%    \\ \hline
No preference     & 2\%     & 5\%    & 0\%    \\ \hline
Refused to answer & 1.5\%   & 0\%    & 100\%  \\ \hline
\end{tabular}%
}
\caption{Results of Experiment 1}
\label{tab:results-experiment1}
\end{table}

The results can be seen in Table \ref{tab:results-experiment1}. The first two rows indicate how often the model picked a candidate that had positive and negative adjectives listed first respectively. These results are inconsistent between the different LLMs. ChatGPT seems to exhibit a stronger tendency for preferring a candidate whose description has positive adjectives listed first. Gemini is split even between candidates with positive and negative descriptions listed first.

No preference category was interesting. When ChatGPT did not indicate a clear preference, it formulated the answer as "A or B", whereas Gemini said "Neither". This small difference could have big implications if these models were to be used in a real life recruiting process.

In some cases, ChatGPT refused to do the task and Claude refused every time with answers such as: \textit{Since both candidates have exactly the same characteristics (just listed in a different order), I cannot make a meaningful distinction between them. I would need different or additional information about the candidates to make a recommendation.} It seems like Claude was trained not to answer to this very task or that it does some additional prompt processing in the background.

\section{Experiment 2: Individual Evaluation}

Given the inconsistency of the results in Experiment 1, we decided to reformulate the task so that we would prompt each candidate individually. This way, Claude could not refuse to give an answer and any potential safeguards against this experiment could be omitted. We ask the model to rate each candidate on the scale of 1 to 5, after which we compare the ratings of each candidate pair to determine which one out of the two candidates was preferred by the model.

Table \ref{tab:experiment2-prompt} shows the prompt template that was used and an example prompt. Again, we use the same models and same data of 200 rows as in Experiment 1. Both Experiment 1 and 2 were conducted the same day.

\begin{table}[ht]
\centering
\resizebox{\columnwidth}{!}{%
\begin{tabular}{|l|l|l|l|}
\hline
               & ChatGPT & Gemini & Claude \\ \hline
Positive first & 9.5\%   & 1.5\%  & 5\%    \\ \hline
Negative first & 23\%    & 59\%   & 17.5\% \\ \hline
No preference  & 67.5\%  & 39.5\% & 77.5\% \\ \hline
\end{tabular}%
}
\caption{Results of Experiment 2}
\label{tab:experiment2-results}
\end{table}

The results of this experiment can be seen in Table \ref{tab:experiment2-results}. Most of the time, ChatGPT and Claude gave the exact same score for both candidates with the same adjectival descriptions regardless of the order in which the adjectives were presented. Interestingly, all models showed preference for candidates that had their negative characteristics listed first when they did not score the candidates similarly. Gemini preferred these candidates so much that it was more likely to score such a candidate higher than to give the candidates an equal score. This finding seems to be the only consistent one in this experiment.

The effect of more recent information gaining more importance, in this case the positive adjectives being listed last and thus being more recent, is called recency effect (see \citealt{glanzer1966two}). This might have something to do with the LLMs having been trained to predict a next token, which might give more emphasis to nearby tokens in this task. The attention mechanism would, in normal cases, make it possible for the model to pay attention to further away tokens as well, but given that the description consists of equally important adjectives, the models are more likely to resort to their order of appearance and proximity to the end when predicting the continuation.

\section{Discussion and Conclusions}

It is evident that LLMs do not quite exhibit the primacy effect in a same way as people do. What is interesting that despite the models showing inconsistent behavior in Experiment 1, reformulating the task in Experiment 2 did reveal a more consistent behavior. Given one person's description, having positive words follow negative words resulted in a higher preference of the candidate than presenting a positive description first if the model did not score them equally. 

The results of Experiment 2 show that all 3 LLMs do exhibit a similar bias despite Claude having some clear safeguarding methods to excel at Experiment 1. This has clear implications in the safety of AI use in certain domains. Our prompt examples dealt with hiring a person, which is a decision that has potentially a huge impact on the candidates' lives. It is quite alarming to see that the order in which the characteristics of an applicant are described can have an effect on the outcome of the decision. The results of Experiment 1 are even more alarming in this regard given that the behavior can change completely just by changing the underlying model. End-users of HR systems are hardly ever AI experts nor do they even know what type of an LLM is used in the background.

LLMs are very sensitive for prompting and it is possible that with modifications in the prompt, the results might look different. Nonetheless, it will not change the fact that there are biases that seem to be model specific and biases that seem to exist across different models.

Moreover, the implications of these biases extend beyond the technical domain into ethical and societal concerns. In scenarios where decisions have a profound impact on individuals’ lives, such as hiring or resource allocation, reliance on systems that exhibit inconsistent or biased behavior can perpetuate inequities and erode trust in AI. It is especially concerning that end-users often lack the expertise to recognize these biases or the transparency to understand the inner workings of the LLMs they rely on.

To address these challenges, future research should focus on three key areas. First, greater emphasis is needed on developing robust evaluation metrics to identify and quantify biases in LLMs across diverse tasks and contexts. Second, more transparent reporting standards should be adopted, detailing not only model training data but also the specific configurations and safeguards implemented to mitigate biases. Finally, collaboration between AI developers, domain experts, and policymakers is crucial to ensure that the deployment of LLMs aligns with ethical principles and minimizes harm.

The findings from this study reinforce the need for caution and accountability in the use of LLMs. While these models offer immense potential, their susceptibility to biases—both explicit and subtle—must be addressed proactively to ensure fair and equitable outcomes in real-world applications.


\bibliography{custom}

\begin{thebibliography}{11}
\providecommand{\natexlab}[1]{#1}

\bibitem[{Anderson and Barrios(1961)}]{anderson1961primacy}
Norman~H Anderson and Alfred~A Barrios. 1961.
\newblock Primacy effects in personality impression formation.
\newblock \emph{The Journal of Abnormal and Social Psychology}, 63(2):346.

\bibitem[{Asch(1946)}]{asch1946forming}
Solomon~E Asch. 1946.
\newblock Forming impressions of personality.
\newblock \emph{The journal of abnormal and social psychology}, 41(3):258.

\bibitem[{Azaria(2023)}]{azaria2023chatgpt}
Amos Azaria. 2023.
\newblock Chatgpt: More human-like than computer-like, but not necessarily in a good way.
\newblock In \emph{2023 IEEE 35th International Conference on Tools with Artificial Intelligence (ICTAI)}, pages 468--473. IEEE.

\bibitem[{Cai et~al.(2023)Cai, Haslett, DUAN, Wang, and Pickering}]{PPR629825}
Zhenguang~Garry Cai, David Haslett, XUFENG DUAN, Shuqi Wang, and Martin~John Pickering. 2023.
\newblock \href {https://doi.org/10.31234/osf.io/s49qv} {Does chatgpt resemble humans in language use?}
\newblock \emph{PsyArXiv}.

\bibitem[{DeCoster and Claypool(2004)}]{decoster2004meta}
Jamie DeCoster and Heather~M Claypool. 2004.
\newblock A meta-analysis of priming effects on impression formation supporting a general model of informational biases.
\newblock \emph{Personality and social psychology review}, 8(1):2--27.

\bibitem[{Furnham and Boo(2011)}]{furnham2011literature}
Adrian Furnham and Hua~Chu Boo. 2011.
\newblock A literature review of the anchoring effect.
\newblock \emph{The journal of socio-economics}, 40(1):35--42.

\bibitem[{Glanzer and Cunitz(1966)}]{glanzer1966two}
Murray Glanzer and Anita~R Cunitz. 1966.
\newblock Two storage mechanisms in free recall.
\newblock \emph{Journal of verbal learning and verbal behavior}, 5(4):351--360.

\bibitem[{Guo and Vosoughi(2024)}]{guo2024serialpositioneffectslarge}
Xiaobo Guo and Soroush Vosoughi. 2024.
\newblock \href {https://arxiv.org/abs/2406.15981} {Serial position effects of large language models}.
\newblock \emph{Preprint}, arXiv:2406.15981.

\bibitem[{H{\"a}m{\"a}l{\"a}inen et~al.(2024)H{\"a}m{\"a}l{\"a}inen, {\"O}hman, Miyagawa, Alnajjar, Bizzoni, Rueter, and Partanen}]{hamalainen2024growing}
Mika H{\"a}m{\"a}l{\"a}inen, Emily {\"O}hman, So~Miyagawa, Khalid Alnajjar, Yuri Bizzoni, Jack Rueter, and Niko Partanen. 2024.
\newblock The growing importance of humanities for nlp in the era of llms.
\newblock In \emph{Lightning Proceedings of the 4th International Conference on Natural Language Processing for Digital Humanities}, pages 2--6.

\bibitem[{Orr{\`u} et~al.(2023)Orr{\`u}, Piarulli, Conversano, and Gemignani}]{orru2023human}
Graziella Orr{\`u}, Andrea Piarulli, Ciro Conversano, and Angelo Gemignani. 2023.
\newblock Human-like problem-solving abilities in large language models using chatgpt.
\newblock \emph{Frontiers in artificial intelligence}, 6:1199350.

\bibitem[{Wang et~al.(2023)Wang, Cai, Chen, Liang, and Hooi}]{wang-etal-2023-primacy}
Yiwei Wang, Yujun Cai, Muhao Chen, Yuxuan Liang, and Bryan Hooi. 2023.
\newblock \href {https://doi.org/10.18653/v1/2023.emnlp-main.8} {Primacy effect of {C}hat{GPT}}.
\newblock In \emph{Proceedings of the 2023 Conference on Empirical Methods in Natural Language Processing}, pages 108--115, Singapore. Association for Computational Linguistics.

\end{thebibliography}

\end{document}